\documentclass[10pt,twocolumn,letterpaper]{article}

\usepackage{cvpr}

\usepackage{booktabs}
\usepackage{subcaption}
\usepackage{multirow}
\usepackage[english]{babel}
\usepackage{microtype}
\usepackage{dcolumn}
\newcolumntype{d}[1]{D{.}{.}{#1}}
\usepackage{pifont}

\definecolor{cvprblue}{rgb}{0.21,0.49,0.74}
\definecolor{green2}{RGB}{0,160,0}
\usepackage[breaklinks,colorlinks,allcolors=cvprblue]{hyperref}

\def\paperID{*****}
\def\confName{CVPR}
\def\confYear{2026}

\title{Materialistic RIR:\\ Material Conditioned Realistic RIR Generation}

\author{
Mahnoor Fatima Saad$^{1}$ \qquad
Sagnik Majumder$^{2}$ \qquad
Kristen Grauman$^{2}$ \qquad
Ziad Al-Halah$^{1}$\\[0.5em]
$^{1}$University of Utah \qquad
$^{2}$UT Austin\\
}
\begin{document}
\maketitle

\begin{abstract}

Rings like gold, thuds like wood! The sound we hear in a scene is shaped not only by the spatial layout of the environment but also by the materials of the objects and surfaces within it.
For instance, a room with wooden walls will produce a different acoustic experience from a room with the same spatial layout but concrete walls.
Accurately modeling these effects is essential for applications such as virtual reality, robotics, architectural design, and audio engineering.
Yet, existing methods for acoustic modeling often entangle spatial and material influences in correlated representations, which limits user control and reduces the realism of the generated acoustics.
In this work, we present a novel approach for material-controlled Room Impulse Response (RIR) generation that explicitly disentangles the effects of spatial and material cues in a scene.
Our approach models the RIR using two modules: a spatial module that captures the influence of the spatial layout of the scene, and a material module that modulates this spatial RIR according to a user-specified material configuration.
This explicitly disentangled design allows users to easily modify the material configuration of a scene and observe its impact on acoustics without altering the spatial structure or scene content.
Our model provides significant improvements over prior approaches on both acoustic-based metrics (up to +16\% on RTE) and material-based metrics (up to +70\%).
Furthermore, through a human perceptual study, we demonstrate the improved realism and material sensitivity of our model compared to the strongest baselines.

\end{abstract}

\section{Introduction}\label{intro}

\begin{figure}[t]
    \centering
    \includegraphics[width=0.9\linewidth]{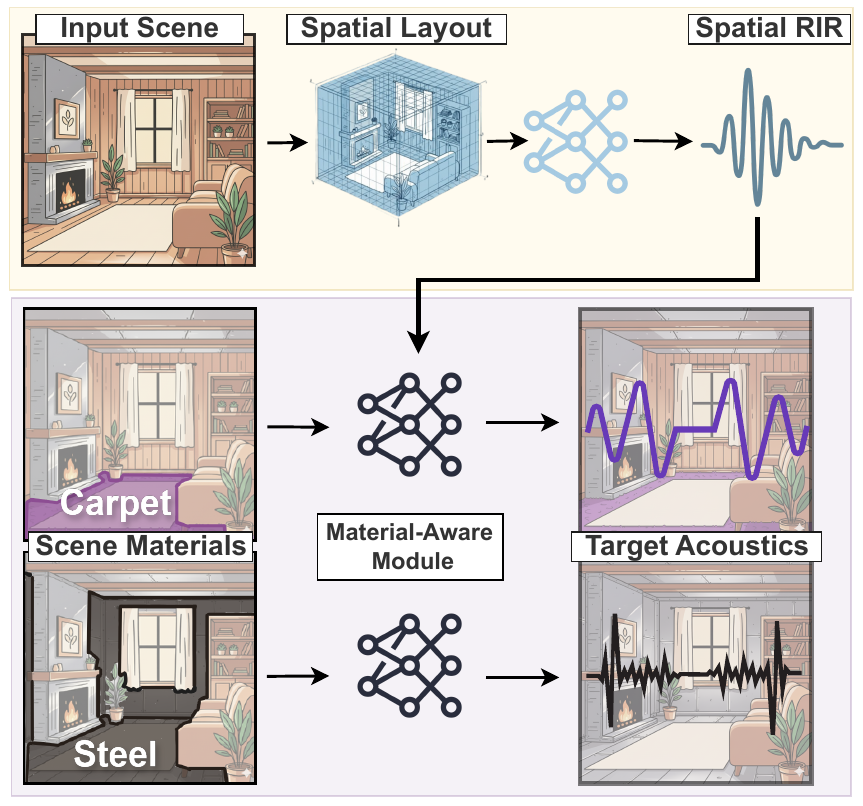}
    \vspace{-0.2cm}
    \caption{        Our explicit disentangled modeling of spatial and material contributions to indoor acoustics (RIRs) enables fine-grained user control over the generated RIR while preserving high acoustic fidelity.
        For example, a user can easily explore how the acoustics of a room, and how music sounds within it, would change by adding a thick carpet to the floor or covering the walls with steel panels.}
    \vspace{-0.2in}
    \label{fig:intro}
\end{figure}

Sound plays a fundamental role in how we perceive and interact with our surroundings.
Beyond conveying speech or music, acoustic cues encode rich semantic, spatial, and material properties of a scene~\cite{AV-RIR,schissler2017acoustic,li2018scene,chen22soundspaces2,tang2022gwa}.
These cues complement visual perception, helping us localize sound sources~\cite{Wu_Wu_Ju_Wang_2021,mo2023avsamsegmentmodelmeets,Tian_2018_ECCV}, navigate complex spaces~\cite{Chen_2021_CVPR,chen2024sim2real}, and interpret subtle social and contextual signals~\cite{wollmer2013youtube,Hong2022VisualCA}.
Accurately modeling acoustic behavior is therefore crucial for a wide range of applications, including augmented and virtual reality (AR/VR)~\cite{zhang2017surround,hammershoi2005binaural,kim2019immersive}, robotics~\cite{chen2024sim2real,9197008}, and spatial audio design~\cite{Chen_2025_CVPR,NEURIPS2024_ff1f4141}.
In these systems, realistic reproduction of acoustic characteristics enhances immersion~\cite{werner2021creation} and bridges the perceptual gap between virtual and physical spaces.
For example, the sound in a rendered theater should match the expected reverberation of a real one.
Consequently, the ability to model and control acoustic cues is key to producing natural and perceptually coherent auditory experiences.

The Room Impulse Response (RIR)~\cite{vigran2014building} describes how sound propagates in a scene and captures its distinctive acoustic signature.
It encodes how an emitted sound wave reflects, is absorbed, and scatters across surfaces before reaching the listener, thereby modeling the complex interactions between a scene's material properties and its spatial configuration.
Variations in room shape, object arrangement, and material composition (e.g., walls made of concrete versus brick) imprint unique temporal and spectral patterns on the resulting reverberation.
Once the RIR of a scene is computed, it can be convolved with any audio signal to reproduce how that sound would be perceived in that specific environment.
As such, RIRs are essential for realistic acoustic rendering, enabling the faithful recreation of auditory experiences in both simulated and real-world settings.

Given its fundamental role in realistic sound modeling, generating accurate and perceptually consistent RIRs has attracted increasing research interest~\cite{liu2025hearing,jin2025avdar}.
Modern approaches aim to infer the acoustic profile of a scene by modeling its visual and spatial characteristics from uni- or multimodal observations such as RGB images~\cite{AV-NERF, AV-RIR, naf, image2reverb}, depth maps~\cite{NACF, AV-RIR}, or audio samples~\cite{Fast-RIR}.
By leveraging these cues, models can estimate how sound interacts with a scene without requiring explicit 3D reconstruction or computationally expensive physical simulation.
While these methods have shown impressive progress, they typically model the scene as a whole by jointly encoding all relevant cues (e.g., visual, spatial, and material) into a single latent representation for RIR prediction.
This joint encoding entangles the contributions of different factors, causing the model to learn implicit correlations between semantics, material properties, and spatial layout.
Hence, users cannot easily control or modify individual aspects of the scene, such as changing wall materials, and get a matching RIR.
This lack of disentanglement limits both flexibility and interpretability, making it difficult to isolate and manipulate specific acoustic attributes or to simulate hypothetical material configurations in a controlled and meaningful way.

Recently, M-CAPA~\cite{saad2025materialrir} made a significant step toward addressing these challenges by introducing a novel dataset and a model capable of generating RIRs for arbitrary material configurations within the same scene.
This approach showed that conditioning on user-defined materials enables meaningful control over the acoustic outcome.
However, although M-CAPA attempts to disentangle material and spatial cues through the design of its training data, it models these factors jointly when generating the target RIR.
As a result, the learned representations remain correlated, which limits fine-grained control.

To address this limitation, we propose a novel model that explicitly disentangles spatial and material cues during RIR generation (see Fig. \ref{fig:intro}).
This disentangled representation allows users to condition the output RIR on arbitrary material configurations while preserving the underlying spatial structure, providing fine-grained and perceptually meaningful control over the resulting acoustics.
In addition, we introduce new evaluation metrics that capture material-dependent acoustic signatures in the RIRs.
These metrics measure how well the predicted responses preserve the spectral and temporal characteristics associated with different materials, complementing standard RIR evaluation metrics.
Through extensive experiments, our model achieves state-of-the-art performance, improving over the best existing baseline by up to 16.8\% and 71.2\% on RTE and material consistency metrics, respectively.
Our results demonstrate superior realism, controllability, and generalization to unseen scenes and material configurations.
We further evaluate the perceptual quality of our model through a user study in which participants rank their preferences between audio generated by our method and by the strongest baselines.
The study shows that our material-conditioned predictions are preferred over SOTA models 60.4\% of the time, further highlighting the realism enabled by our disentangled approach to RIR generation.

\section{Related Works}
\label{sec:related_works}

Room Impulse Response (RIR) estimation is an important problem with applications in domains such as audio-visual navigation~\cite{chen20soundspaces, Chen_2021_CVPR, 9197008, majumder2021move2hear, majumder2022active}, localization~\cite{Wu_Wu_Ju_Wang_2021,mo2023avsamsegmentmodelmeets,Tian_2018_ECCV}, speech enhancement~\cite{AV-RIR, Adverb,ratnarajah2023improvedroomimpulseresponse}, and augmented and virtual reality (AR/VR)~\cite{liu2021soundsynthesispropagationrendering,kim2019immersive}. Since the literature spans a broad range of settings and applications, we focus here on the two directions most relevant to our work: learning-based RIR estimation from multimodal observations, and modeling material properties and their impact on acoustics.

\vspace{-.4cm}
\paragraph{Multimodal RIR Generation}
Traditional RIR estimation methods rely on physics-based wave propagation models or audio-visual simulators~\cite{FDTD, 10.1121/1.2164987, 10.1121/1.3021297, 5165582, MEHRA201283,10.1145/2980179.2982431, 10.1121/1.398336, 10.1121/1.382599,5753892,9414399}.
While effective, these approaches typically require extensive manual measurements or detailed \textit{a priori} knowledge of the environment, which limits their ability to generalize to unseen scenes.
To reduce this dependence on explicit scene information such as full 3D meshes or geometry~\cite{Holters2009IMPULSERM, stan2002comparison}, later work explored learning-based alternatives from partial multimodal observations.
Early methods focused on estimating late reverberation from scene images~\cite{kon2018deep,kon2020auditory} or room geometry~\cite{Fast-RIR,remaggi2019reproducing,inras}.
More recent approaches generate full RIRs from visual inputs~\cite{image2reverb}, use scene-specific audio-visual observations to model RIRs for particular scenes or novel viewpoints~\cite{chen2022visual,somayazulu2024self,li2024self,chen2023novel}, or incorporate audio-visual-text cues in simulated environments~\cite{chen2025meanrirmultimodalenvironmentawarenetwork}.
Other multimodal methods estimate acoustics at arbitrary locations using sampled RIRs and images~\cite{majumder2022fsrir,NACF}, or scene geometry and poses~\cite{Fast-RIR,naf,inras,hearinganythinganywhere2024}.
However, despite the use of rich multimodal cues, these methods often overlook the role of material properties in shaping acoustics, which limits the realism of the generated RIRs.
In contrast, our approach explicitly models material properties and their effect on RIR generation, enabling fine-grained control over material configurations and their acoustic impact.

\vspace{-.4cm}
\paragraph{Material Conditioned RIR Estimation}
The acoustic properties of different materials, as well as their distribution within a space, strongly influence the acoustics of an environment.
Some methods have therefore explicitly modeled materials for RIR generation~\cite{AV-RIR,Listen2Scene,schissler2017acoustic,li2018scene,Liang_2025_ICCV,gao2024soaf,av_gs}, and have shown clear improvements over material-agnostic approaches~\cite{AV-RIR}.
However, these methods either require explicit material annotations for semantic objects in a scene~\cite{AV-RIR,Listen2Scene} or depend on full 3D mesh reconstruction to estimate materials~\cite{schissler2016interactive,li2018scene,tang2022gwa,Listen2Scene,jin2025avdar}.
As a result, they have limited ability to estimate accurate RIRs under novel material configurations.
More recently, M-CAPA~\cite{saad2025materialrir} addressed this challenge by estimating RIRs conditioned on arbitrary material configurations from a single audio-visual observation.
However, although it explicitly models materials, it does not disentangle spatial and material cues during RIR generation, which leads to learning correlated representations.
In addition, its reliance on both audio and visual observations reduces its practicality in real-world settings, where obtaining clean audio samples is often difficult.
In contrast, we propose a vision-only approach that generates material-conditioned RIRs while explicitly modeling both material acoustics and spatial cues of the scene, enabling generating of higher quality RIRs with greater control over material impact on acoustics.

\section{Material Conditioned RIR Generation}\label{sec:task}

This work addresses the material-conditioned RIR generation task, where the goal is to predict RIRs that faithfully reflect custom object and surface material compositions in a scene~\cite{saad2025materialrir}.
More specifically, given an RGB image of a 3D scene and desired material assignments, we aim to estimate the RIR that matches both the spatial scene cues from the image, and the specified object materials.
Here, we consider an RIR with both the source and receiver located at the image capture position, and is obtained by emitting a short frequency sweep and recording the resulting echoes.

Formally, we are given \textbf{1)} an RGB image $V$ of size $H \times W$, captured with a 90$^{\circ}$ field-of-view (FoV) camera; \textbf{2)} a material segmentation mask $M$ of size $H \times W$, where each index denotes one of $N$ material classes that can be assigned to objects and surfaces; and \textbf{3)} a two-channel (binaural) audio spectrogram $A$ with size $2 \times T \times F$ corresponding to the RIR recorded at the camera location, which reflects the spatial cues in $V$ and the material assignments in $M$.
Here, $T$ and $F$ denote the numbers of temporal windows and frequency bins, respectively, and $A$ is computed by applying a short-time Fourier transform (STFT) to the corresponding binaural RIR.
Our objective is to train a model $\mathcal{F}$ such that
$\mathcal{F}(V, M) = \hat{A},$
where $\hat{A}$ is an estimate of the target RIR spectrogram $A$.

This task is particularly challenging because the material assignments $M$ is user-specified and can be arbitrary.
As a result, the same scene represented by $V$ may correspond to different material assignments, leading to subtle changes in the RIR, and the model must generate accurate RIRs even for material configurations not observed during training.

\label{sec:model}
\begin{figure*}[th!]
    \centering
    \includegraphics[width=0.8\linewidth] {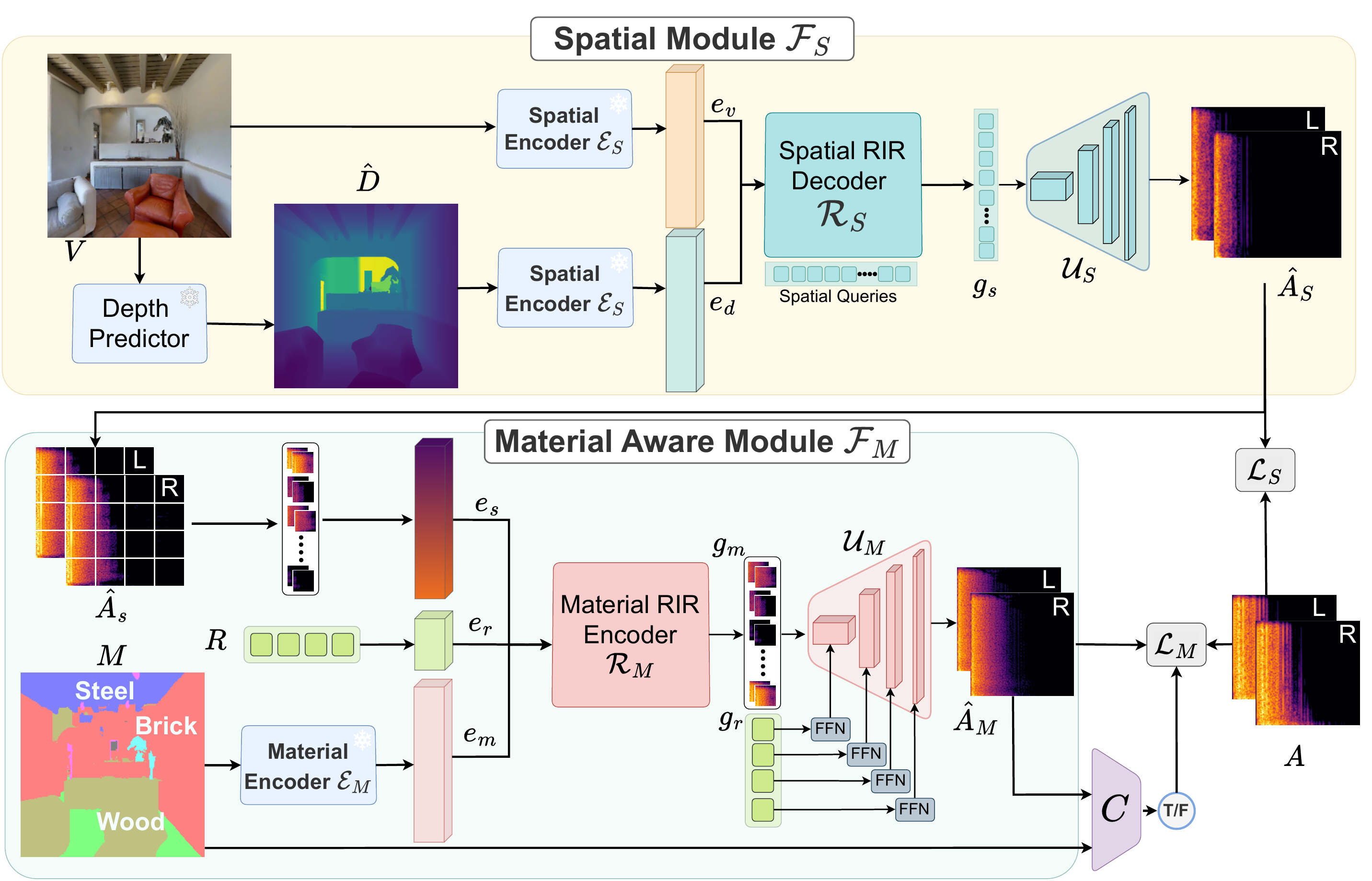}
    \caption{        Our MatRIR model $\mathcal{F}$ for material-conditioned RIR prediction.
        Given an RGB image $V$ of a 3D scene, MatRIR uses the Spatial Module $\mathcal{F}_S$ to extract geometric cues from $V$ and predict a spatially accurate initial estimate of the target RIR, $\hat{A}_S$.
        Then, for a material segmentation mask $M$, which specifies a custom object material configuration for $V$, the Material-Aware Module $\mathcal{F}_M$ modulates $\hat{A}_S$ to produce the final RIR estimate, $\hat{A}_M$.
        We train the model with a combination of losses that enforces acoustic consistency between ground-truth RIR $A$ and $\hat{A}_M$ and promotes cross-modal correspondence between $\hat{A}_M$ and $M$, which substantially improves performance.}
        \label{fig:model}
        \vspace{-0.4cm}
    \end{figure*}

\section{MatRIR Model}\label{sec:model}

We propose the Material-Aware RIR Network (MatRIR), which separately models the spatial and material acoustic effects of a scene, leading to high-fidelity RIR generation.

Our MatRIR consists of two acoustic estimation modules (see Fig.~\ref{fig:model}):
1) a \textbf{Spatial Module $\mathcal{F}_S$}, which predicts RIRs from inputs that only encode the spatial layout of a scene, such as depth maps, thereby capturing how the arrangement of walls and objects shapes scene acoustics; and
2) a \textbf{Material-Aware Module $\mathcal{F}_M$}, which modulates the initial spatial RIR estimate from $\mathcal{F}_S$ to account for the effects of different surface material assignments.

This factorized design yields high-quality RIR predictions that more faithfully capture the effect of material assignments on scene acoustics, as shown in our results.
In the following, we describe the architecture and training of MatRIR in detail.

\subsection{Spatial Module}\label{sec-spatial}
The goal of the Spatial Module $\mathcal{F}_S$ is to model how the spatial layout of a scene shapes its acoustics and to use this information to generate spatially accurate RIRs.
$\mathcal{F}_S$ consists of a Spatial Encoder $\mathcal{E}_S$, a Spatial RIR Decoder $\mathcal{R}_S$, and an Audio Feature Upsampling Network $\mathcal{U}_S$ (Fig.~\ref{fig:model} top).

Given an input image $V$, $\mathcal{F}_S$ first uses the encoder $\mathcal{E}_S$ to extract the scene's spatial cues.
These spatial features are then passed to the RIR decoder $\mathcal{R}_S$, which uses learned queries to cross-attend to the features and learn a representation that captures the spatial attributes of the target RIR.
Finally, the decoder output is upsampled by the audio feature upsampler $\mathcal{U}_S$ to produce an initial estimate of the target RIR, $\hat{A}_S$.
Next, we describe these components in detail.

\vspace{-0.1in}
\paragraph{Spatial Encoder $\mathcal{E}_S$}
Given the image $V$, we first use a pretrained depth predictor (MiDaS~\cite{birkl2023midas}) to obtain a depth map estimate $\hat{D}$, where $\hat{D} \in [0, 1]^{H \times W}$, $H$ and $W$ denote the image height and width, respectively, and each pixel stores the normalized depth at that location.
We then use a pretrained image encoder (DINOv2-Large~\cite{oquab2023dinov2}) to embed the image $V$ and depth map $\hat{D}$ into visual features $e_v$ and depth features $e_d$, respectively.
Specifically, we extract the outputs of the 18$^{\text{th}}$ layer of the DINOv2 encoder, yielding  $e_v$ and $e_d$ embeddings such that each modality is represented by 256 tokens with dimension 1024.
RGB and depth provide complementary spatial information.
While depth provides coarse cues about the spatial layout of a scene, RGB captures finer details about the arrangement of objects and surfaces, which can benefit spatial RIR estimation.

\vspace{-0.3cm}
\paragraph{Spatial RIR Decoder $\mathcal{R}_S$}
Given the spatial encoding of the scene captured by the visual $e_v$ and depth $e_d$ features, the Spatial RIR Decoder $\mathcal{R}_S$
augments these features with  learned modality-specific embeddings $s_v$ and $s_d$~\cite{majumder2022fsrir}.
First,  $\mathcal{R}_S$ combines $e_v$ with $s_v$ and $e_d$ with $s_d$, and then fuse the resulting features using modality-specific projections. This enables $\mathcal{R}_S$ to capture both cross-modal and modality-specific environmental cues.
The enriched features are concatenated into a single feature sequence $f \in \mathbb{R}^{256 \times 512}$.
Then, $\mathcal{R}_S$ uses a 4-layer transformer decoder that employs a set of learned spatial queries~\cite{10.5555/3618408.3619222} to cross-attend to the fused features $f$ and extract the features $g_s$ that captures the spatial properties of the target RIR.

\vspace{-0.3cm}
\paragraph{Audio Feature Upsampling Network $\mathcal{U}_S$}
Given the spatial features $g_s$, we first reshape them into 2D feature maps.
We then feed the reshaped features into the upsampler CNN $\mathcal{U}_S$, which uses a stack of four transposed convolution layers to upsample the features and produce the initial estimate of the target RIR, $\hat{A}_S$, based on the spatial cues of the scene.

\subsection{Material-Aware RIR Module}
Our Material-Aware RIR Module $\mathcal{F}_M$ is designed to capture how surface and objects materials influence scene acoustics, enabling its RIR predictions to model material-dependent phenomena such as sound transmission, absorption, and reflection.
Specifically, $\mathcal{F}_M$ takes the initial RIR estimate $\hat{A}_S$ from the Spatial Module $\mathcal{F}_S$ and modulates it with material-dependent acoustic attributes based on the material segmentation mask $M$ (cf.~Sec.~\ref{sec:task}).

$\mathcal{F}_M$ consists of a Material Mask Encoder $\mathcal{E}_M$, a Material RIR Encoder $\mathcal{R}_M$, and a Material-Aware Audio Feature Upsampler $\mathcal{U}_M$.
Given the material mask $M$, $\mathcal{F}_M$ first uses the encoder $\mathcal{E}_M$ to extract material features.
It then feeds these features to the RIR encoder $\mathcal{R}_M$ together with spatial audio features extracted from $\hat{A}_S$.
$\mathcal{R}_M$ extracts material cues from the material features and uses them to modulate the spatial audio features through self-attention, producing a feature sequence that captures both spatial and material information.
Finally, a CNN $\mathcal{U}_M$ upsamples the output of $\mathcal{R}_M$ to obtain a material-conditioned estimate of the target RIR, $\hat{A}_M$.
Next, we describe each component in detail.

\vspace{-0.3cm}
\paragraph{Material Encoder $\mathcal{E}_M$}
Similar to the Spatial Encoder $\mathcal{E}_S$, the Material Encoder $\mathcal{E}_M$ uses a pretrained DINOv2-Large encoder~\cite{oquab2023dinov2} to embed the material segmentation mask $M$ into a sequence of features $e_m$, where $e_m \in \mathbb{R}^{256 \times 1024}$.
The features in $e_m$ are extracted from the 18$^{\text{th}}$ layer of the encoder.
The resulting material features are then passed to the Material RIR Encoder $\mathcal{R}_M$ for further processing.

\vspace{-0.1in}
\paragraph{Material RIR Encoder $\mathcal{R}_M$}
Given the RIR estimate $\hat{A}_S$ from the Spatial Module $\mathcal{F}_S$, we extract patches from each channel of $\hat{A}_S$ and use a small MLP to encode each binaural patch, producing a sequence of spatial audio features $e_s$.
We additionally introduce a sequence of audio feature re-weighting tokens~\cite{hu2018squeeze}, $R$, and project them into re-weighting features $e_r$.
We then feed $e_m$, $e_r$, and $e_s$ to the RIR encoder $\mathcal{R}_M$, which applies self-attention over all input features and produces material-aware audio features $g_m$ and re-weighting features $g_r$.
The features $g_r$ capture useful spatial and material cues that are later used to modulate~\cite{hu2018squeeze} the importance of different audio features at different resolutions in $\mathcal{U}_M$, which improves RIR prediction quality, as shown in our results (see Table~\ref{tab:ablations}, row~b).
We implement $\mathcal{R}_M$ as a 4-layer transformer encoder~\cite{dosovitskiy2020vit} and use four re-weighting tokens.

\vspace{-0.3cm}
\paragraph{Material-aware Audio Feature Upsampler $\mathcal{U}_M$}
Given the material-aware audio features $g_m$, we first reshape them into 2D feature maps and then use a 4-layer transposed convolution network $\mathcal{U}_M$ to upsample them.
$\mathcal{U}_M$ also takes the re-weighting features $g_r$ as input and uses them to modulate the outputs of each upsampling layer, allowing the model to learn which features are most important for modeling the material conditioned scene acoustics.
We treat the output $\hat{A}_M$ of $\mathcal{U}_M$ as the final estimate that captures both the spatial and material information of the target RIR, ${A}$.

\subsection{Model Training}\label{sec:app_metrics}
We train our model with a combination of losses to encourage fine-grained acoustic similarity between the predicted and target RIRs, while also improving the correspondence between the predicted RIR and the input material mask $M$.
In addition, we minimize the prediction error for both the spatial RIR estimate $\hat{A}_S$ and the material-aware RIR estimate $\hat{A}_M$.
To this end, we define the overall training loss as $\mathcal{L} = \mathcal{L}_S + \mathcal{L}_M$, where $\mathcal{L}_S$ and $\mathcal{L}_M$ denote the losses for $\hat{A}_S$ and $\hat{A}_M$, respectively, such that:
\begin{small}
\begin{gather}
    \mathcal{L}_S = \lambda_1 \|\hat{A}_S - A\|_1 + \lambda_2 L_D(\hat{A}_S, A) \nonumber \\
    \mathcal{L}_M = \lambda_1 \|\hat{A}_M - A\|_1 + \lambda_2 L_D(\hat{A}_M, A) + \lambda_3 L_C(\hat{A}_M, M) \nonumber
\end{gather}
\end{small}
where $\|\cdot\|_1$ denotes the $L_1$ loss between the predicted and ground-truth magnitude spectrograms; $L_D$ is an energy decay loss~\cite{majumder2022fsrir} that encourages the predicted and target RIRs to follow similar energy decay patterns, which helps the model better capture late reverberation; and $L_C$ is a novel auxiliary loss that measures how well the predicted RIR $\hat{A}_M$ corresponds to the material segmentation mask $M$.

To model this cross-modal correspondence through $L_C$, we pretrain a material-RIR matcher network $\mathcal{C}$ to predict 1 when given a matching RIR and material mask, and 0 otherwise.
During training of our model $\mathcal{F}$, we freeze the weights of $\mathcal{C}$, feed it the material segmentation mask $M$ and the predicted RIR $\hat{A}_M$, and use its output as an estimate of how well $\hat{A}_M$ matches $M$.
We use the backpropagated error through $\mathcal{C}$ to provide material-conditioned feedback to $\mathcal{F}_M$, thereby improving the material-awareness of our predicted RIRs.
Adding $L_C$ to the training objective provides strong material-conditioned supervision for RIR prediction and improves performance, as shown in our results (Table~\ref{tab:ablations}, row~c).
See Supp for more details.

\subsection{Material Evaluation Metrics}\label{sec:material_eval_metrics}
We introduce two evaluation metrics to quantify how effectively an RIR generation model captures the key acoustic effects of scene materials.
As we show in Sec.~\ref{sec:eval_main}, standard RIR metrics do not capture this important aspect well.
Next we provide an overview of these metrics and provide additional details and evaluation setup in Supp.

\vspace{-0.2cm}
\paragraph{Material Classification Accuracy (MatC).}
This metric measures how effectively a model encodes the acoustic properties of a \emph{single material} in the generated RIR.
To compute this metric, we pretrain an RIR-based material classifier on a simplified setup in which all objects in a scene are assigned the same material.
We then apply this classifier to the generated RIRs and measure its accuracy.
A model that correctly captures material-dependent acoustic effects should yield high material classification accuracy under this metric.

\vspace{-0.2cm}
\paragraph{Material Distribution Accuracy (MatD).}
This metric measures how effectively a model captures the scene's \emph{distribution of materials} in the generated RIR.
To learn this metric, we first perform k-means clustering~\cite{9072123} on the material distributions derived from the material masks in the training dataset.
Each cluster corresponds to samples with similar material distributions.
We then train a classifier to predict, from an RIR, the cluster label associated with its sample, thereby capturing the material composition of the scene that is encoded in the input RIR.
To evaluate an RIR prediction model, we classify its predicted RIRs $\hat{A}$ into one of 36 clusters and report the top-5 accuracy.
This metric provides a measure of how well the predicted RIR reflects the material distribution of the objects in the scene.

While MatC and MatD are not direct measures of RIR quality, they provide important complementary insights into how well a model captures material-dependent acoustic effects in its RIR predictions, which is a key aspect of material-conditioned RIR generation.
In particular, MatC considers a single-material setting and thus provides a more direct measure of how well a model captures the acoustic properties of individual materials that should be reflected in the generated RIR.
In contrast, MatD evaluates performance in a more realistic and challenging setting in which multiple materials are present, requiring the model to capture the overall material distribution in the RIR.
To ensure a reliable evaluation, the networks used to compute MatC and MatD are trained only on the training scenes of the dataset and never access the test data or test scenes used for evaluating any model.

\section{Experiments}
\label{sec:experiments}

\begin{table*}[!t]
    \centering
    \tiny
    \resizebox{1.\textwidth}{!}{
    \setlength{\tabcolsep}{3pt}
    \renewcommand{\arraystretch}{1.}
    \begin{tabular}{l|r r r r|r r|r r r r|r r|r r r r|r r}
    \toprule
        & \multicolumn{6}{c|}{\textbf{Seen Materials} $D_{us}$} & \multicolumn{6}{c|}{\textbf{Unseen Materials}  $D_{uu}$} & \multicolumn{6}{c}{\textbf{Unseen Pairings}  $D_{uk}$} \\
    \textbf{Method}
    & \multicolumn{1}{c}{\textbf{L1}} & \multicolumn{1}{c}{\textbf{STFT}} & \multicolumn{1}{c}{\textbf{RTE}} & \multicolumn{1}{c|}{\textbf{CTE}}& \multicolumn{1}{c}{\textbf{MatC}}& \multicolumn{1}{c|}{\textbf{MatD}} & \multicolumn{1}{c}{\textbf{L1}} & \multicolumn{1}{c}{\textbf{STFT}} & \multicolumn{1}{c}{\textbf{RTE}} & \multicolumn{1}{c|}{\textbf{CTE}} & \multicolumn{1}{c}{\textbf{MatC}}& \multicolumn{1}{c|}{\textbf{MatD}} & \multicolumn{1}{c}{\textbf{L1}} & \multicolumn{1}{c}{\textbf{STFT}} & \multicolumn{1}{c}{\textbf{RTE}} & \multicolumn{1}{c|}{\textbf{CTE}} & \multicolumn{1}{c}{\textbf{MatC}} & \multicolumn{1}{c}{\textbf{MatD}}\\
    \midrule

     {Image2Reverb~\cite{image2reverb}} & {14.68} & {7.89} & {245.2} & {18.76}&10.01& 9.01& {14.13} & {7.59} & {223.3}  & {19.15} &9.33&9.19& {14.98} & {8.19} & {244.5} & {19.55}&9.87&9.23 \\

     {FAST-RIR++~\cite{Fast-RIR,majumder2022fsrir}}  & {16.73} & {25.06} & {317.2} & {21.47} &9.20&11.20&{14.81} & {28.39} & {231.8} & {16.83} &9.10&13.0& {16.41} & {31.02}  & {321.0} & {21.18}&9.13&13.60 \\

    JM-CNN    & 5.89 & 5.34 & 185.7  & 9.42  & 10.88 & 12.40 & 6.10&5.63 &192.1  &9.98 &10.83&10.55& 6.31  &6.09 &191.6  & 10.05 &10.80&12.38 \\

    JM-Transformer    & 6.72 & 8.85  & 134.8  & 12.22  & 9.76 &11.00 & 6.90 &9.04 & 139.5  &12.56  &9.36&8.05&7.14  &9.61 &136.4  & 12.38 &9.82& 11.09\\

    JM-QFormer     & 6.09 & 6.17  & 101.3  & 11.05  & 18.10 &10.20&6.23 &6.44 &98.63  &11.49 &18.09&8.75&6.49  &6.97 &102.0  &11.67 &18.13&12.43 \\

    M-CAPA~\cite{saad2025materialrir}  & 5.92 & 5.49 & 89.23 & \textbf{8.41}&9.32&21.85&6.06  & 5.76 &92.80 &\textbf{ 9.05} &9.75&20.65&6.30& 6.17& 91.75& 8.95&9.55&23.42 \\

    \midrule
     MatRIR (Ours)     & \textbf{5.46}  & \textbf{5.21}  & \textbf{75.56}  & 8.53  & \textbf{89.26} &\textbf{ 31.75}&\textbf{ 5.60} &\textbf{ 5.41} & \textbf{77.18} & 9.16 &\textbf{89.29} &\textbf{31.01}  & \textbf{5.85} & \textbf{5.84} &\textbf{77.69} &\textbf{8.74}&\textbf{89.13}&\textbf{30.91} \\

    \bottomrule
    \end{tabular}
    }
    \caption{        Results on unseen environments for our three test splits: $D_{us}$ with seen material profiles, $D_{uu}$ with unseen material profiles, and $D_{uk}$ with unseen profile pairings. STFT and $L_1$ are scaled by $\times 10^{-2}$, RTE is in milliseconds (ms), and CTE is in decibels (dB). Lower values indicate better performance for acoustic metrics. MatC and MatD are percentage accuracy, higher is better for these metrics.}
    \vspace{-0.2cm}
    \label{tab:main}
\end{table*}

We evaluate our approach on the Acoustic Wonderland dataset~\cite{saad2025materialrir} and compare it with several state-of-the-art methods~\cite{saad2025materialrir, Fast-RIR, image2reverb} and strong baselines (Sec.~\ref{sec:eval_main}) to assess the accuracy of the generated RIRs.
We further provide detailed model analysis and ablations in Sec.~\ref{sec:ablations}, as well as a user study comparing the quality of the produced RIRs with the strongest baselines (Sec.~\ref{sec:user_study}).

\vspace{-0.4cm}
\paragraph{Implementation Details}
Our approach takes a $256\times256$ RGB image $V$ as input, together with a target material mask $M$ for the scene.
To represent the material properties of scene objects, we use a material segmentation map in which each element of $M$ is the index of one of the $N$ material classes defined in~\cite{saad2025materialrir}.
Our goal is to predict a 0.5\,s, 16\,kHz binaural room impulse response (RIR) conditioned on the material configuration of the scene.
We represent binaural RIRs as spectrograms computed using the Short Time Fourier Transform (STFT).
We use a 16ms Hanning window ~\cite{1455106}, and 2ms hop length, which results in two-channel spectrograms of size $256\times256$.
We train the model on a single GPU using the Adam optimizer~\cite{adam-optimizer} with cosine annealing learning rate scheduling~\cite{loshchilov2016sgdr}, an initial learning rate of $7\times10^{-5}$, and a batch size of 150.

\vspace{-0.4cm}
\paragraph{Acoustic Wonderland Dataset (AcoW)}
We use the same data splits and evaluation protocol as in~\cite{saad2025materialrir}.
In AcoW, 76 scenes are designated as \textit{seen} ($|\mathcal{S}_s| = 76$), and 8 as \textit{unseen} ($|\mathcal{S}_u| = 8$).
Three scenes from $\mathcal{S}_u$ are used for validation and the remaining five for testing.
The 2673 material configurations are divided into \textit{seen configurations}, with $|\mathcal{C}_s| = 2405$, and \textit{unseen configurations}, with $|\mathcal{C}_u| = 268$.
The training set, $D^{tr} = \{\mathcal{S}_s, \mathcal{C}_s\}$, contains 1.28M samples.
The evaluation set has three splits each containing 2000 samples: $D_{us} = \{\mathcal{S}_u, \mathcal{C}_s\}$, $D_{uu} = \{\mathcal{S}_u, \mathcal{C}_u\}$, and $D_{uk} = \{\mathcal{S}_u, \mathcal{C}_P\}$ where $\mathcal{C}_P$ represents a set of unseen material configuration pairings (see~\cite{saad2025materialrir} for details).
This results in a total test set of 6000 samples.

\vspace{-0.4cm}
\paragraph{SoTA Methods and Baselines}\label{sec:baselines}
We compare our approach against the following SoTA methods and baselines (see Supp for more details):
\begin{itemize}[leftmargin=*,align=left]

    \item \textbf{Image2Reverb}~\cite{image2reverb} is a vision-only SoTA model that predicts RIR using RGB and depth images of a scene.
    \item \textbf{FAST-RIR++}~\cite{Fast-RIR,majumder2022fsrir} is a vision-only GAN-based model for RIR prediction, with improvements proposed by~\cite{majumder2022fsrir}.
    \item \textbf{M-CAPA~\cite{saad2025materialrir}} is a SoTA multimodal, material-conditioned approach to RIR estimation that uses the material features to condition the prediction of the RIR. For fair comparison, we consider the vision-only version of M-CAPA that uses RGB as input, similar to our model.
    \item \textbf{JM-CNN} is a baseline that jointly model all scene cues using the same input features as our model ($V, \hat{D},M$) but model these features jointly with a CNN decoder to predict the target RIR.
    \item \textbf{JM-Transformer} is similar to above, but uses a transformer encoder similar to our material RIR encoder (Figure \ref{fig:model}) to encode the interaction between the input features before using a CNN decoder to predict the RIR.
    \item \textbf{JM-QFormer} follows a network design similar to our Spatial RIR Decoder (Figure \ref{fig:model}), where the transformer decoder learns a set of tokens, by cross attending against $V, \hat{D}, M$ before predicting the RIR.
\end{itemize}
These sets of models allow comparison with SoTA vision-based RIR generation methods, while the JM-* baselines explore whether different joint modeling designs of scene features can achieve better material-controlled RIR generation in contrast to our approach that shares some of these design elements but explicitly model the spatial and material cues separately and their respective impacts on target RIR.

\vspace{-0.2cm}
\paragraph{Evaluation Metrics}
We evaluate performance using standard RIR metrics:
1) \textbf{$L_1$ Distance}, which measures the $L_1$ distance between the predicted and ground-truth magnitude spectrograms of the RIRs;
2) \textbf{STFT Error}, which measures the mean squared error between the previous spectrograms;
3) \textbf{RT60 Error (RTE)}~\cite{ratnarajah2021ts}, which measures the error in the RT60 value of the predicted RIR; and
4) \textbf{Early-to-Late Index Error (CTE)}~\cite{ratnarajah2021ts}, which measures the error in the ratio of early-to-late received energy.
Additionally, we report the proposed material-aware evaluation metrics, 5) \textbf{MatC} and 6) \textbf{MatD}, introduced in Sec.~\ref{sec:material_eval_metrics}.

\subsection{Material-Controlled RIR Generation Results}\label{sec:eval_main}
Table~\ref{tab:main} compares the performance of our MatRIR model with SOTA and baseline methods across the three test splits.
We observe that our joint modeling baselines, which are trained to predict material-conditioned RIRs from all input modalities $V$, $\hat{D}$, and $M$, outperform Image2Reverb~\cite{image2reverb} and FAST-RIR++~\cite{Fast-RIR, majumder2022fsrir}, which only model the spatial layout of the scene.
This highlights the importance of accounting for material distribution when predicting accurate RIRs.
Among prior methods, M-CAPA (Vision)~\cite{saad2025materialrir} outperforms all baselines and vision-only state-of-the-art approaches on the standard metrics and on MatD.

Our MatRIR further outperforms all baselines and prior methods on most standard metrics across all splits, demonstrating the effectiveness of our approach.
More importantly, MatRIR not only produces more accurate RIRs, but also better captures the key aspects of the scene material configuration, outperforming all baselines on MatC and MatD by 71.12\% and 9.25\% on average, respectively.
These results show that explicitly separating spatial and material properties in RIR modeling improves performance while also enabling more direct and fine-grained control over material assignments.
Finally, we observe that improvements in the standard metrics do not always correlate with similar trends in the material-based metrics.
This further motivates the need for the proposed material-sensitive metrics, MatC and MatD, to better understand the performance of different models.

\subsection{Perceptual Material User Study}\label{sec:user_study}
We conduct a user study to test the perceptual quality of the material-dependent acoustic effects produced by our model.
Participants listened to audio samples generated by M-CAPA and MatRIR and selected the sample that sounded more realistic for a given target material mapping $\mathcal{M}_T$.
The study involved 7 participants, each of whom evaluated 22 samples.
Across all responses, 60.4\% favored MatRIR as producing the more realistic audio.
These results show that MatRIR not only captures material-dependent acoustics more accurately, but also yields higher perceptual realism, producing RIRs that better reflect the assigned scene materials.

\begin{table}[!t]
    \centering
    \resizebox{1.\columnwidth}{!}{
    \setlength{\tabcolsep}{2pt}
    \renewcommand{\arraystretch}{1.}
    \begin{tabular}{l|r r r r|rr}
    \toprule
        \textbf{Method} & \textbf{L1} & \textbf{STFT} & \textbf{RTE} & \textbf{CTE} &\textbf{MatC} &\textbf{MatD} \\
    \midrule
    MatRIR (Ours)           & 5.60 &5.41  &77.18 &9.16 &89.29&31.0 \\
    \midrule
    a) w/o $\mathcal{C}$  &  5.44	&5.23	&78.94&	8.34	&65.02&	29.30 \\
    b) w/o  $R$  &  6.49&	7.13&	142.4&	8.98	&20.02&	11.20 \\
    c) w/ $(V, D)$ Only  &  6.06&	5.71&	154.7	&9.99	&9.09	&9.95 \\
    d) w/ $M$ Only  &5.74	&5.58&	97.78	&8.86&	18.20	&17.25 \\
    \bottomrule
    \end{tabular}
    }
    \caption{Ablation of our model on the split $D_{uu}$. Lower is better for L1, STFT, RTE and CTE; higher is better for MatC and MatD}
    \vspace{-0.5cm}
    \label{tab:ablations}
\end{table}
\begin{figure}[t]
    \centering
        \includegraphics[width=0.82\columnwidth]{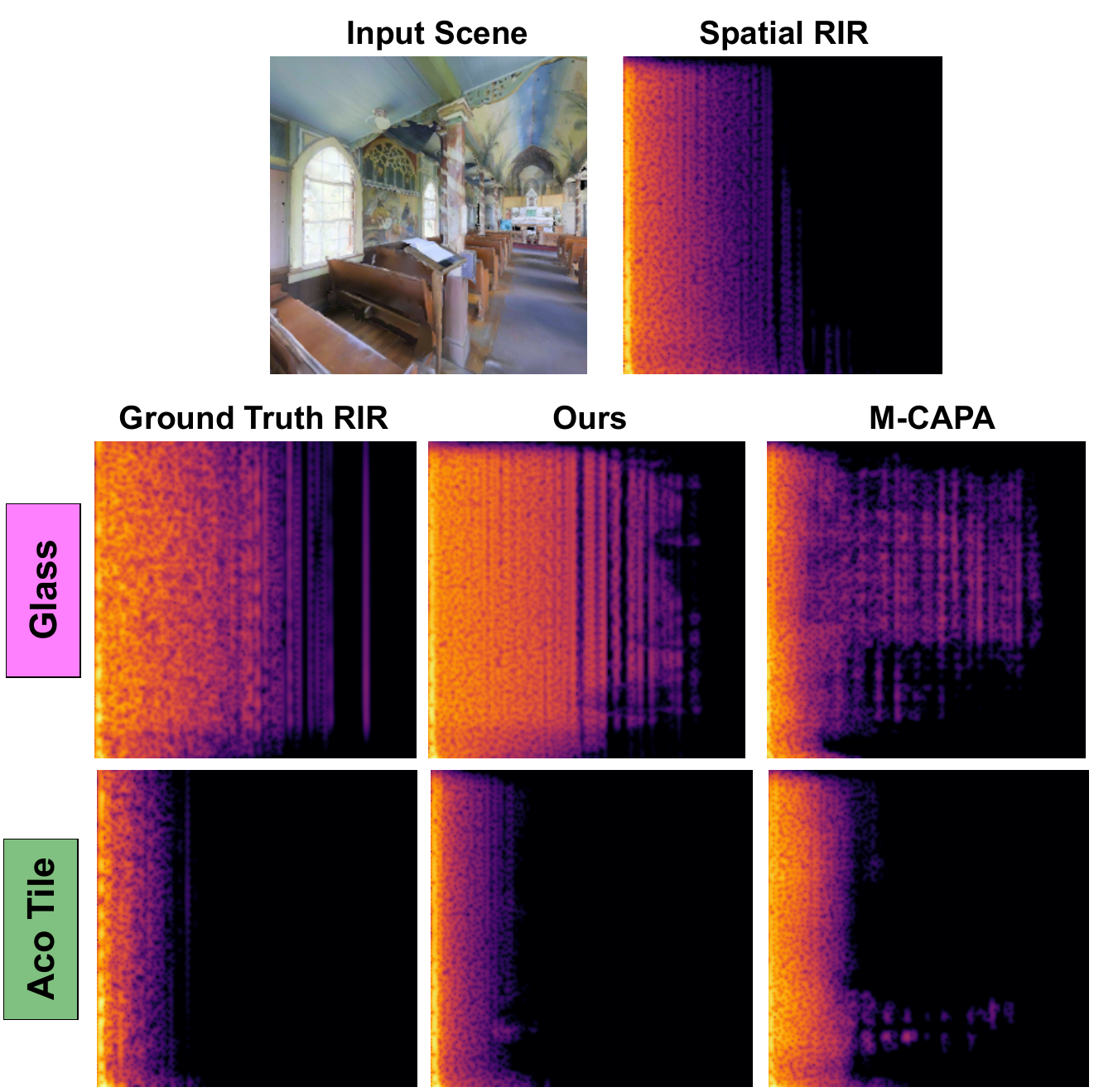}
    \caption{Qualitative results where everything in the scene is assigned to a single material. Unlike M-CAPA, our model is able to accurately capture each material's unique acoustic properties.}
    \label{fig:all_same}
    \vspace{-0.5cm}
\end{figure}
\begin{figure*}[th]
    \centering
    \includegraphics[width=0.8\linewidth] {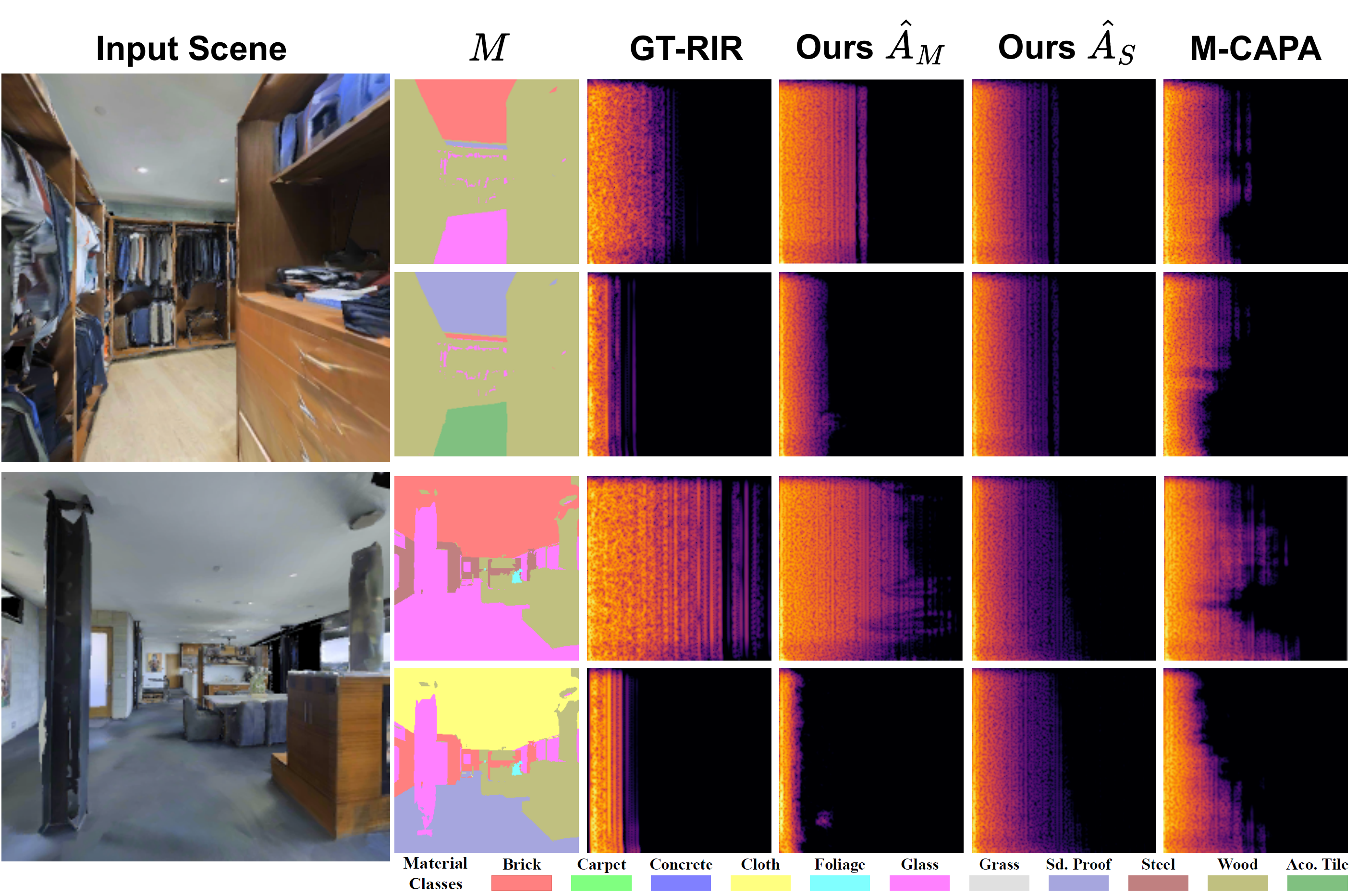}
    \caption{Qualitative results of MatRIR.
    Our model generates spatially accurate RIRs $\hat{A}_S$ and modulates them to capture material-specific acoustic properties.
    Given the same scene with different material configurations $M$, MatRIR generates the corresponding material-dependent patterns in the predicted RIRs $\hat{A}_M$ more faithfully than the competing method, M-CAPA.
    For visualization, we show only one channel of the binaural RIR predictions.}
    \label{fig:qual2}
    \vspace{-.1cm}
\end{figure*}

\subsection{Model Analysis}\label{sec:ablations}
To better understand the behavior of our model, we provide next a set of ablations and in-depth analysis of the results.

\vspace{-0.3cm}
\paragraph{Ablations}
Table~\ref{tab:ablations} shows the contribution of each component of MatRIR.
Removing the material matcher (row~a) or disabling the injection of material cues at different stages of the final RIR prediction (row~b) leads to substantial drops in MatC and MatD.
We further evaluate the two modules of MatRIR separately (rows~c and~d).
Using only the Spatial Module output $\hat{A}_S$ results in poor performance across all metrics, especially MatC and MatD, indicating that spatial cues alone are insufficient.
On the other hand, using only the Material Module without spatial layout cues (i.e., w/o $\hat{A}_S$) also yields poor performance, showing that material information alone is not enough to accurately predict an RIR.
These results show that both spatial and material cues are necessary, and that modeling them separately and explicitly, as in MatRIR, is more effective.
See Supp for an analysis of performance across different material classes and for a comparison of computational cost and parameter count between our model and the baselines.

\vspace{-0.3cm}
\paragraph{Qualitative Results}
Figure~\ref{fig:qual2} shows qualitative comparisons between our model and M-CAPA~\cite{saad2025materialrir}.
For the same scene, we present two different material configurations sampled from the test set $D_{uu}$.
Notably, the spatial RIR prediction $\hat{A}_S$ remains unchanged across material configurations because it is conditioned only on the spatial layout of the scene.
In contrast, the final prediction $\hat{A}_M$ is modulated by the input material mask and changes accordingly, even when the material change affects only a relatively small region of the scene (top example).
The same behavior is observed when the material configuration changes more substantially (bottom example).
By comparison, M-CAPA shows limited sensitivity to small material changes and responds more clearly only to larger ones.
Even in those cases, however, it fails to match the target RIR as accurately as our model.

\begin{figure}[t]
    \centering
        \includegraphics[width=0.82\columnwidth]{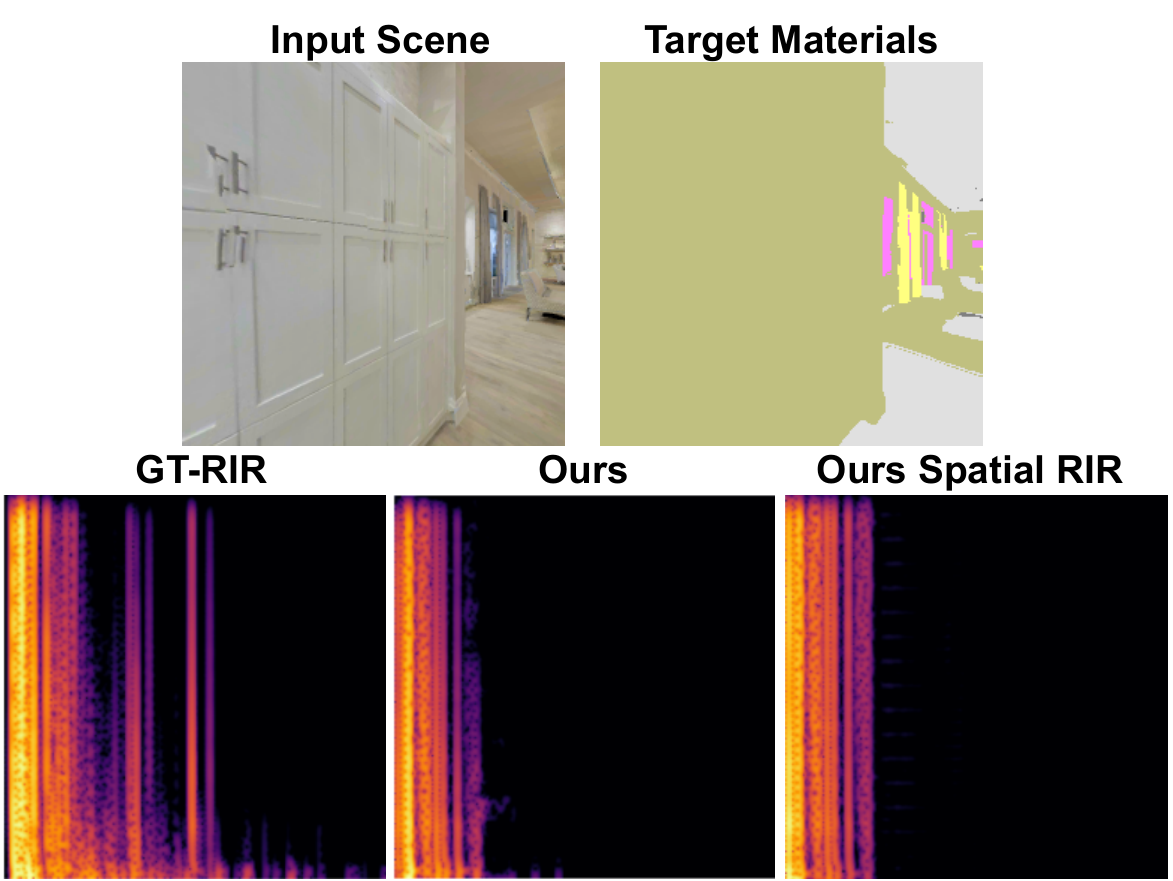}
    \caption{For scenes where the space is not fully visible, for example, the room hidden by the wall of cabinets here, our model struggles to capture the material distribution adequately and relies on the spatial aspects more than the material signal.}
    \label{fig:failure}
    \vspace{-0.5cm}
\end{figure}

We further illustrate this behavior in cases where the entire scene is assigned a single material (Fig.~\ref{fig:all_same}).
In this setting, M-CAPA produces different RIRs for large changes in the input material configuration, but does not clearly capture the distinct acoustic signatures of different materials.
In contrast, our model better captures these material-specific characteristics.

\vspace{-0.3cm}
\paragraph{Failure Cases and Limitations}
One failure case of our model is shown in Fig.~\ref{fig:failure}.
We observe such failures when the input image is captured too close to a wall.
In these cases, the model tends to rely more heavily on spatial cues and produces an RIR that is less sensitive to the material configuration.
We expect that this happens because the limited field of view provides insufficient information about the full scene layout and surrounding materials, making it harder for the model to apply the correct material-based modulation to the spatial RIR.
A possible way to address this limitation is to incorporate a full $360^\circ$ view of the scene, which we leave for future work.

\section{Conclusion}
In this work, we introduced a novel approach for realistic RIR generation that disentangles spatial and material cues from the environment and models them with separate components, enabling more explicit control over acoustic properties.
As a result, our approach produces more realistic, flexible, and perceptually accurate RIRs, achieves state-of-the-art performance among vision-only baselines, and shows clear improvements in both acoustic metrics and user preference.

\newpage

{
    \small
    \bibliographystyle{ieeenat_fullname}
    \bibliography{main}
}

\newpage
\clearpage

\setcounter{page}{1}
\section{Supplementary Material}
\label{sec:supp}

In this supplementary material, we provide further analyses and details about our approach:

\begin{itemize}
    \item Sec.~\ref{sec:supp_video}: Supplementary video (with audio) for qualitative results.
    \item Sec.~\ref{sec:further_analysis}: Additional analysis of our model for fine-grained changes to the material mask.
    \item Sec.~\ref{sec:quals_supp}: Real-world experiments.
    \item Sec. \ref{sec:supp_user_study}: Details about the user study on perceptual evaluation of our model.
    \item Sec.~\ref{sec:supp_limits}: Limitations and discussion.
    \item Sec.~\ref{sec:comp_costs}: Computation cost analysis of our model and the baselines.
    \item Sec.~\ref{sec:matc_matd_setup}: Details of the proposed evaluation metrics for measuring our predictions' ability to capture the material-dependence of scene acoustics as mentioned in Sec.~\ref{sec:material_eval_metrics}.
    \item Sec.~\ref{sec:model_architecture}: Model architecture.
    \item Sec.~\ref{sec:supp_eval_setup}: Evaluation setup and baselines .
\end{itemize}

\subsection{Supplementary Video} \label{sec:supp_video}

We provide a supplementary video with audio to qualitatively demonstrate our model outputs (see the project page).
Specifically, we present speech samples from the LibriSpeech~\cite{librispeech} dataset convolved with the RIRs predicted by our model, and compare them with samples convolved using the outputs of the best-performing baselines.
The results highlight the superior ability of our approach to capture fine-grained changes in \emph{both} the scene materials and spatial layout.

\subsection{Additional Model Analyses}\label{sec:further_analysis}
In this section, we further analyze our model's ability to capture fine-grained variations in spatial layout and material configuration.
We also study how the dominant material in the input material mask affects model performance.

\begin{figure}[t]
    \centering
    \begin{subfigure}[t]{1.\columnwidth}
        \centering
        \includegraphics[width=\linewidth]{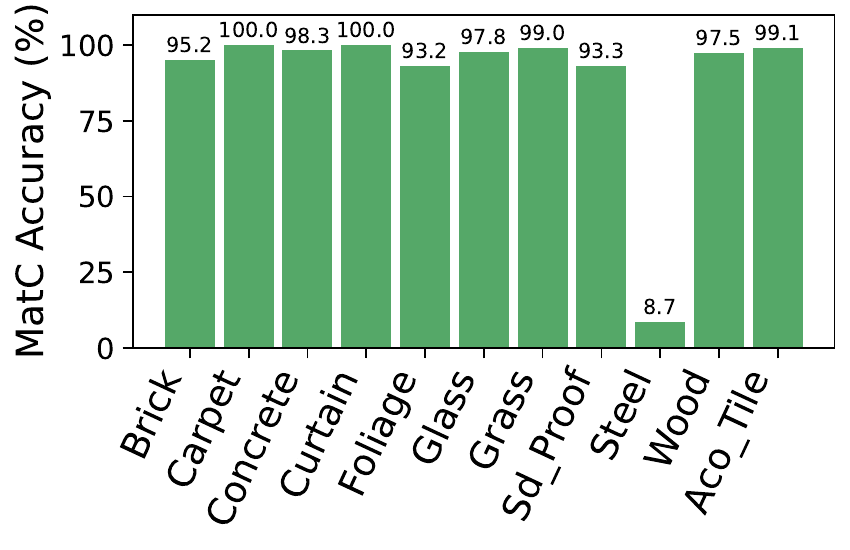}
        \caption{MatC accuracy vs. material class}
        \label{fig:acc_mat_duu}
    \end{subfigure}
    \\
    \vspace{0.3cm}
    \begin{subfigure}[t]{1.\columnwidth}
        \centering
        \includegraphics[width=\linewidth]{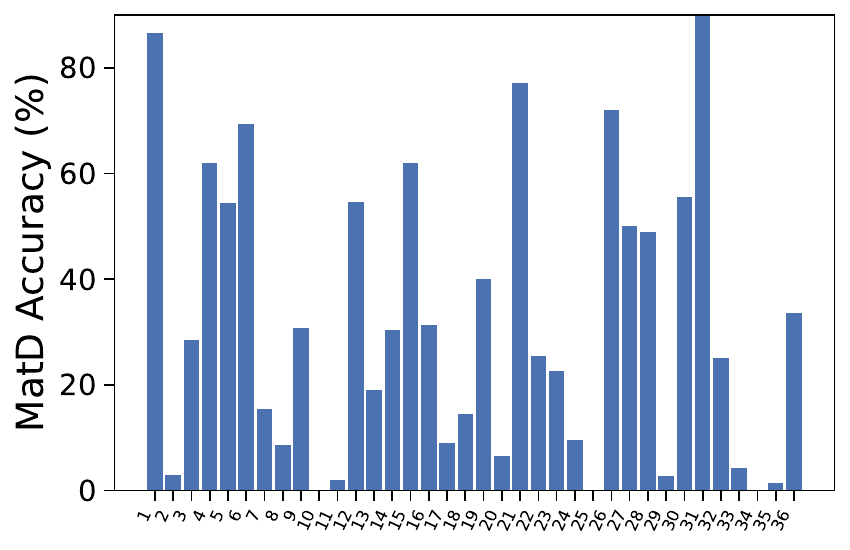}
        \caption{MatD accuracy vs. material distribution}
        \label{fig:matd_eval_plot}
    \end{subfigure}
    \caption{        Evaluation of acoustic modeling with respect to the type and distribution of material present in the target material mask.}
        \label{fig:error}
    \end{figure}
\subsubsection{Material Composition Analysis}
Here, we evaluate how difficult each material in the AcoW dataset is for our model to capture in its predicted RIRs.
To this end, we assign a single material to all objects in the input scene and plot in Fig.~\ref{fig:acc_mat_duu} how model performance varies with the assigned material.
Specifically, the figure reports the MatC metric---material classification accuracy (cf.~Sec.~\ref{sec:matc-details})---for each material in the AcoW dataset.
Notably, our model captures the acoustics of most materials accurately; however, \textit{steel} appears to be particularly challenging.

We further evaluate our model's ability to capture mixtures of materials in the predicted RIR using the MatD metric (cf.~Sec.~\ref{sec:matd-details}).
Figure~\ref{fig:matd_eval_plot} shows this accuracy for different material distributions over the 11 material classes in the dataset.
We use the cluster center (see Sec.~\ref{sec:matd-details}) as the representative of each group of similar material distributions.
In total, we consider 36 distinct material distributions.
Our model captures the acoustic characteristics of scenes with material distributions in clusters 1 and 31 accurately.
By analyzing the dominant materials at these cluster centers, we find that cluster 1 contains a large proportion of \textit{sound-proof} material (44\%) and \textit{glass} (24\%), while cluster 31 contains a large proportion of \textit{steel} (43\%) and \textit{glass} (16\%).
These results indicate that our model can accurately capture the acoustic properties of such material mixtures.

In contrast, our model performs poorly on cluster 10, which contains a mixture of \textit{carpet} (28\%) and \textit{concrete} (29\%).
We observe similarly low performance on cluster 25, with 28\% \textit{acoustic tile} and 27\% \textit{grass}, and on cluster 34, with 28\% \textit{concrete} and 27\% \textit{brick}.
These material mixtures appear to be challenging for our model to capture acoustically, resulting in 0\% MatD accuracy for these distributions.

\subsubsection{Fine-Grained Environmental Changes}
Our approach to material-conditioned RIR generation explicitly disentangles the spatial and material components of the room impulse response.
This factorized modeling enables fine-grained and interpretable control, allowing us to modify scene materials while preserving the underlying geometry.
As a result, our method can generate accurate and meaningful variations in environmental acoustics conditioned on different material properties.

\begin{figure*}
    \centering
    \includegraphics[width=1.\linewidth]{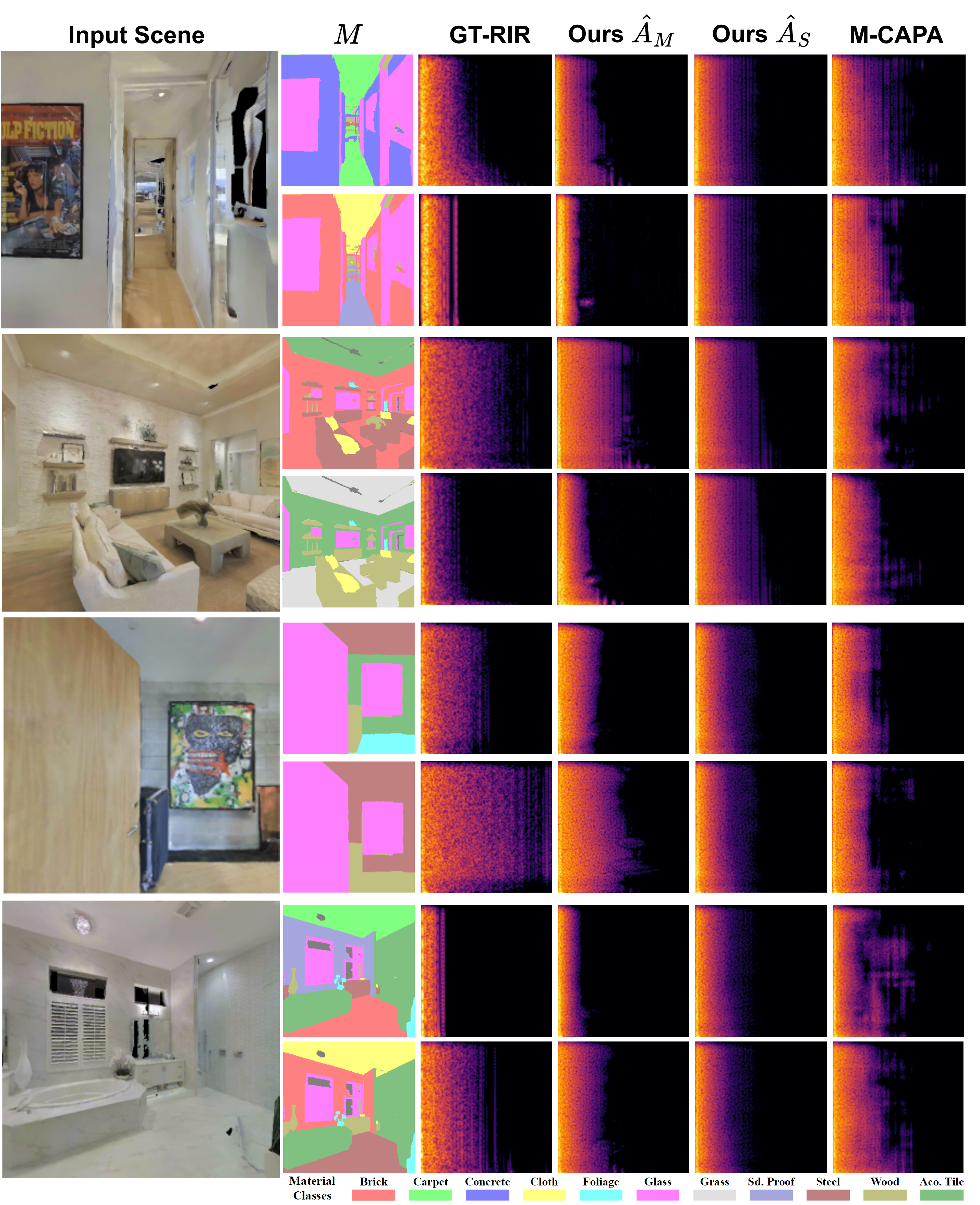}
    \caption{Sample outputs of our model in 4 different scenes.
    Our model is able to accurately capture acoustic changes ($\hat{A}_M$) conditioned on the given materials' configuration $M$.
    For brevity, we only show one channel of the binaural RIR.}
    \label{fig:supp-qual}
\end{figure*}

To illustrate this, Fig.~\ref{fig:supp-qual} shows qualitative examples demonstrating the fine-grained material control provided by our model.
For each example, we use two distinct material configurations while keeping the spatial layout fixed.
The predicted RIRs show that our model consistently preserves the spatial characteristics of the ground-truth RIR while incorporating the target material properties in $M$.
In particular, our approach reliably modulates the spatial RIR $\hat{A}_S$ so that the resulting reverberation patterns reflect the material configuration of the environment, producing RIRs that are consistently closer to the ground truth.
In contrast, M-CAPA provides limited control over changes in scene materials and often generates nearly identical RIRs across different material configurations.
For example, in the first scene, when the wall material changes from concrete to brick, our model modulates the spatial RIR $\hat{A}_S$ to accurately predict the material-conditioned RIR $\hat{A}_M$, whereas the predictions of M-CAPA remain largely unchanged across the two settings.

\begin{figure}
    \centering
    \includegraphics[width=0.95\columnwidth]{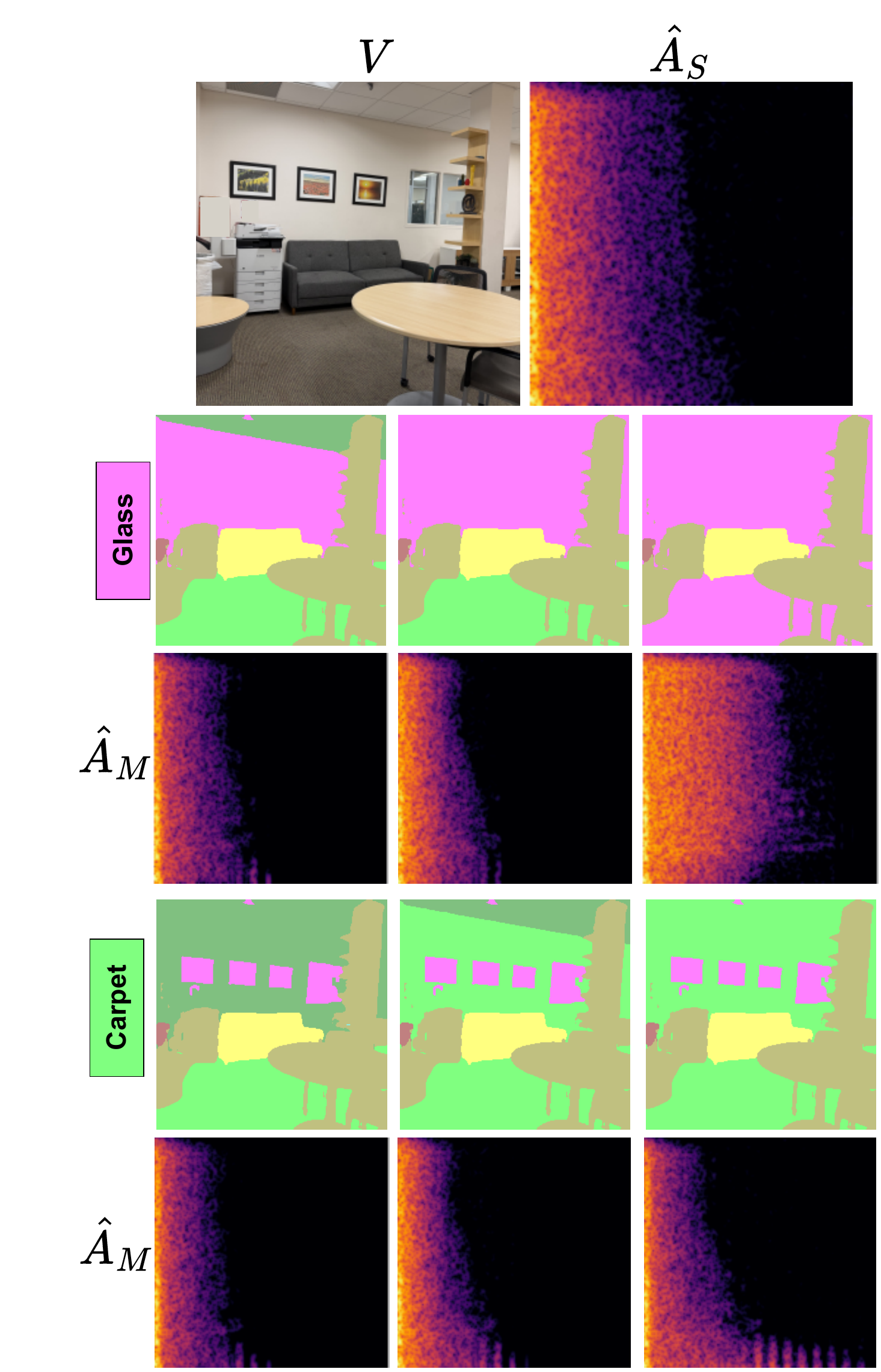}
    \caption{Qualitative examples of our model predictions in a real-world setup.
    We experiment with different material configurations for the same scene by gradually increasing the proportion of a single material class in the target material mask---\emph{glass} (top) and \emph{carpet} (bottom)---and show the resulting material-conditioned RIRs, $\hat{A}_M$, produced by our model.}
    \label{fig:real-world-mat}
\end{figure}

\subsection{Real-World Experiments}\label{sec:quals_supp}
To validate the applicability of our approach beyond simulated environments, we qualitatively evaluate our model on real-world scenes.
To this end, we collect snapshots from different locations inside a building and run inference on these images with our model.
We first extract depth maps using MiDaS~\cite{birkl2023midas} and semantic segmentation masks using InternImage~\cite{wang2022internimage}.
We then use the RGB and depth map as inputs to obtain the spatially accurate initial RIR prediction, $\hat{A}_S$.
From the semantic mask, we generate material masks by assigning objects to material classes, which are then used to produce the material-aware final prediction, $\hat{A}_M$.

In Fig.~\ref{fig:real-world-mat}, we show example predictions for a single RGB view, while varying the target material mask.
By changing the object-to-material assignments in the scene, we demonstrate how our model modulates its initial estimate, $\hat{A}_S$, to reflect different material configurations.
For example, assigning \textit{glass} or \textit{carpet} to the floor, walls, and ceiling produces distinct patterns in the final RIR prediction, $\hat{A}_M$, indicating that changes in scene materials meaningfully affect the modeled acoustics.
These qualitative results show that our model can produce material-aware RIRs in real-world settings.

\begin{figure}
    \centering
    \includegraphics[width=1.0\columnwidth]{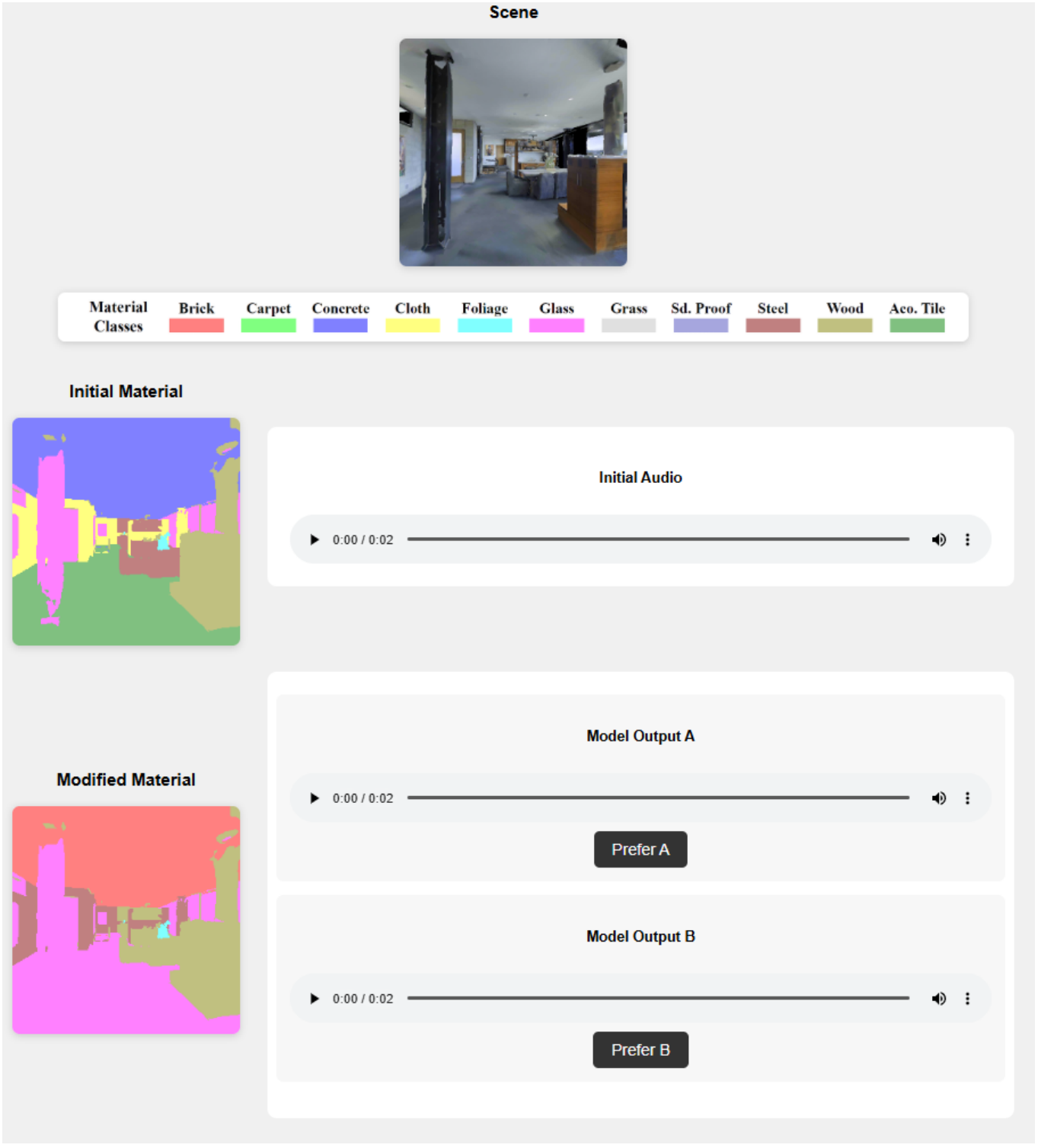}
    \caption{Interface for our user study where participants listen to predicted RIRs from MatRIR and M-CAPA, convolved with clean speech audio, and select which model produces more realistic acoustics.}
    \label{fig:user_study_interface}
\end{figure}

\subsection{Perceptual Material User Study}\label{sec:supp_user_study}
To evaluate the perceptual quality of our model predictions, we conduct a user study in which participants rank two outputs: one from our approach and one from the state-of-the-art baseline, M-CAPA~\cite{saad2025materialrir}.
Figure~\ref{fig:user_study_interface} shows the interface used in the study.

For this study, we select 22 random samples from our hardest test split, $D_{uu}$, in which both the environments and material configurations are unseen during training.
Importantly, all qualitative examples shown in this paper are drawn from this split.
For each sample, we first convolve the predicted RIRs with clean anechoic speech from the LibriSpeech~\cite{librispeech} dataset.

Participants are then shown an RGB image of the scene and asked to:
\textbf{1)} observe a target material mask and listen to audio convolved with the ground-truth RIR for the corresponding spatial layout and material configuration, to understand how the material properties affect the spatial audio;
\textbf{2)} listen to the spatial audio produced by our method and M-CAPA under a new material configuration but the same spatial layout; and
\textbf{3)} rank the two models based on how well their predictions match the target materials.
We find that participants prefer the predictions of our MatRIR over those of M-CAPA in 60.4\% of cases.
This result indicates that our method better captures changes in the material composition of the environment than prior work.

\begin{figure}[t]
    \centering
    \begin{subfigure}[t]{0.9\columnwidth}
        \centering
        \includegraphics[width=\linewidth]{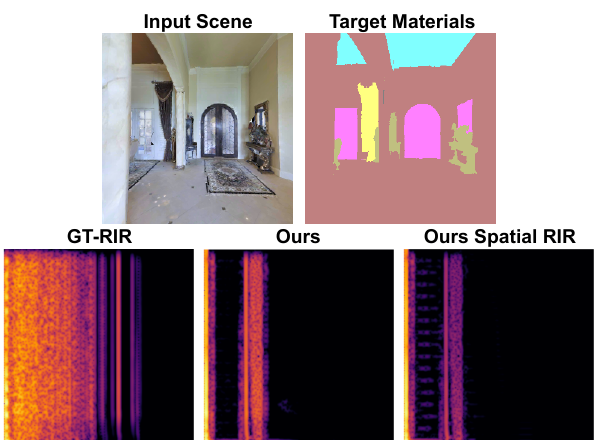}
        \caption{}
        \label{fig:fail2}
    \end{subfigure}
    \\
    \begin{subfigure}[t]{0.9\columnwidth}
        \centering
        \includegraphics[width=\linewidth]{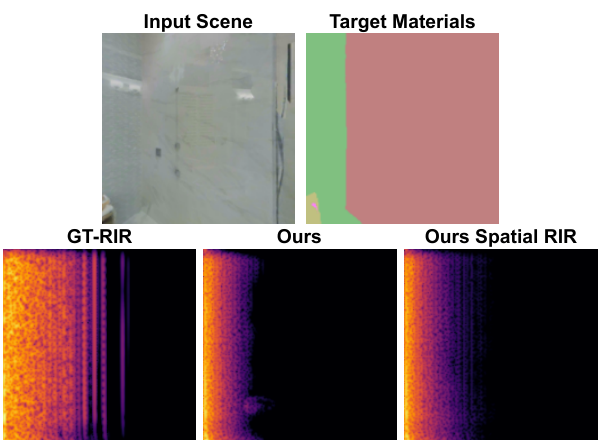}
        \caption{}
        \label{fig:fail3}
    \end{subfigure}
    \vspace{-0.2cm}
    \caption{        Qualitative examples in which our model fails to accurately capture material dependence.
        We observe that when severe occlusions are present and \textit{steel} is the dominant material in the target material mask, our model struggles to capture acoustic effects beyond those already encoded in its geometry-conditioned initial estimate, $\hat{A}_S$, resulting in limited material awareness.}
    \label{fig:supp_failure}
\end{figure}

\subsection{Limitations and Discussion}\label{sec:supp_limits}
We show that our disjoint modeling for RIR prediction quantitatively outperforms strong baselines and prior SOTA on both acoustic and material-based metrics.
Qualitative results further demonstrate the effectiveness of our approach in producing spatially and materially aware RIRs.
However, as shown in Fig.~\ref{fig:failure} in the main paper, our model cannot accurately capture the material distribution of a scene when the scene is only partially visible or when the visual observation is heavily occluded by large objects such as walls or shelves.
In such cases, a possible remedy is to provide a full panoramic observation so that the model can condition on the complete $360^{\circ}$ visual scene.

Furthermore, Fig.~\ref{fig:acc_mat_duu} shows that our model struggles to capture the acoustic properties of \textit{steel} when all objects in the scene are assigned this material.
This observation is consistent with the qualitative examples in Figs.~\ref{fig:fail2} and~\ref{fig:fail3}, where the target material mappings contain a large amount of \textit{steel}.
In these cases, our model fails to accurately capture the corresponding material acoustics.

Generating accurate room acoustics conditioned on the surface materials of objects within a scene remains a challenging problem.
Although our method models the effect of the 11 material classes introduced in~\cite{saad2025materialrir}, expanding the dataset to include a broader range of material representations would help capture the diversity of material distributions encountered in real-world environments.

\begin{table}[t]
    \centering
    \resizebox{1.\columnwidth}{!}{
    \setlength{\tabcolsep}{2pt}
    \renewcommand{\arraystretch}{1.}
    \begin{tabular}{l|r r r}
    \toprule
        \textbf{Method} & \textbf{Params (M)} & \textbf{GFLOPs} & \textbf{Inf. Time (ms)}\\
    \midrule
    Image2Reverb \cite{image2reverb} &  57.6 & 276.91 & 198.44\\
    FAST-RIR\cite{Fast-RIR}++ &  132.68 & 57.84 & 121.76\\
    M-CAPA &  5.84 & 11.24 & 76.61\\
    JM-CNN &   14.99 &10.71& 187.86\\
    JM-Transformer &  6.22 &6.21& 227.81\\
    JM-QFormer &  7.34 & 7.29& 293.10\\
    MatRIR (Ours) & 13.28  &14.11& 270.56\\
    \bottomrule
    \end{tabular}
    }
    \caption{Computational cost of our approach (MatRIR) and the baselines.
    \vspace{-0.5cm}}
    \label{tab:computational_metrics}
\end{table}

\subsection{Computational Costs} \label{sec:comp_costs}
Table~\ref{tab:computational_metrics} compares our model with the baselines in terms of the number of trainable parameters, GFLOPs, and inference time.
When reporting the parameter count, we exclude modules such as DINOv2~\cite{oquab2023dinov2} and MiDaS~\cite{birkl2023midas}, since their weights are frozen during training.

\subsection{Metrics for Evaluating Material Specificity}\label{sec:matc_matd_setup}
In this section, we provide additional details on our proposed evaluation metrics, MatC and MatD, for assessing RIR quality with respect to material awareness.

\subsubsection{Material Classification Accuracy (MatC)}\label{sec:matc-details}
The MatC metric measures how effectively a method captures the impact of materials on RIRs in \emph{single-material setups}.
Specifically, it uses a network trained to predict the material type from a given RIR, where all objects in the scene are assigned the same material.
To compute this metric, we freeze the above-mentioned material classifier, run inference with it and compute its  accuracy vis-a-vis predicting the expected material type as its output.

We pre-train the material classification network using the SSv2~\cite{chen22soundspaces2} simulator and the MP3D~\cite{Matterport3D} dataset.
We use the same splits as M-CAPA~\cite{saad2025materialrir}, with 76, 3, and 5 scenes for training, validation, and testing, respectively.
The total number of ground-truth RIRs is 167K for training, 6.6K for validation, and 11K for testing.

To generate these RIRs, we randomly sample a single material class from the AcoW~\cite{saad2025materialrir} dataset, assign it to all objects in a scene, and capture the echo responses from different locations and viewpoints.
Using this data, we train a ResNet18~\cite{resnet} on 2-channel binaural RIR spectrograms to classify each sample into one of the 11 material classes in AcoW.
During training, we add Gaussian noise to the RIRs to improve robustness to microphone noise~\cite{kehling2024evaluation,inras}.
We train the classifier for 50 epochs using cross-entropy loss~\cite{10.5555/3618408.3619400} with a learning rate of 0.01.
This classifier achieves 96.7\% accuracy on the test set.

\subsubsection{Material Distribution Accuracy}\label{sec:matd-details}
This metric measures how accurately a model captures, in its predicted RIR, the acoustic signature of \emph{a distribution of materials} in the scene.
 To this end, we use a network pretrained to predict the scene's material distribution from an RIR, where, unlike MatC, different objects in the scene can have different materials.

For this metric, we use the same training, validation, and test splits described in Sec.~\ref{sec:experiments}.
To pretrain this network, we first extract the material class distribution from the material segmentation mask $M$ as the percentage of area covered by each material class.
Next, we group these distributions into 36 clusters using K-means~\cite{9072123}, which groups scenes with similar material configurations.
We then train a ResNet18~\cite{resnet} to predict, from a binaural RIR, the cluster index corresponding to the material distribution associated with that input.
This cluster classification network is trained for 10 epochs using cross-entropy loss~\cite{10.5555/3618408.3619400} with a learning rate of $0.001$.
The network achieves 77\% Top-5 accuracy on the test split.
To compute this metric, we freeze the network, run inference on the predicted RIRs, and measure its Top-5 accuracy in predicting the expected material distribution cluster.

\subsection{Model Architecture}\label{sec:model_architecture}
In this section, we provide further details about our model's architecture and training.

\vspace{-0.3cm}
\paragraph{Depth Predictor}
We use a pretrained model from \cite{birkl2023midas} to obtain the depth map input for our model by feeding in the visual observation as an RGB image $V$ and obtaining a depth map.
We normalize this depth map and use it as a grayscale image.

\vspace{-0.3cm}
\paragraph{Spatial Encoder}
Our spatial encoder $\mathcal{E}_S$ is a pretrained, frozen DINOv2-Large \cite{oquab2023dinov2} encoder that takes the visual observation RGB ($V$) and depth map ($D$) and produces feature maps $e_v, e_d \in \mathbb{R}^{256 \times 1024}$ extracted from the 18th layer, where each feature map has 256 tokens and is 1024-dimensional.

\vspace{-0.3cm}
\paragraph{Material Encoder}
Our material encoder $\mathcal{E}_M$ is a pretrained, frozen DINOv2-Large \cite{oquab2023dinov2} encoder that takes a material mapping $M$ as input and produces a feature map $e_m \in \mathbb{R}^{256 \times 1024}$ from the 18th layer, where each of the 256 tokens is represented by 1024 channels.

\vspace{-0.3cm}
\paragraph{Spatial RIR Decoder}
To learn spatial acoustic features, we follow Blip-2~\cite{10.5555/3618408.3619222} architecture for our Spatial RIR Decoder $\mathcal{R}_S$.
We implement a 4-layer transformer decoder with 256 dimensionality, feedforward layers of 512 dimensionality, and a dropout rate of 0.1.
To identify each input modality in this decoder, we follow \cite{majumder2022fsrir} by using modality-specific embeddings $s_v$ for vision modality $V$ and $s_d$ for depth modality $D$.
These modalities are concatenated with their respective modality features $e_v,s_v$ and $e_d,s_d$, and passed through a projection layer to bind modality-specific features.
The final modality-specific feature tokens are concatenated into a single sequence and projected to decoder dimensionality $f \in \mathbb{R}^{512\times256}$ and used as keys and values.
Meanwhile, spatial queries of length 256 are used as the queries for our decoder.
The output from this module $\mathcal{R}_S$ is learned tokens $g_s \in \mathbb{R}^{512\times256}$.

\vspace{-0.3cm}
\paragraph{Material RIR Encoder}
This module, $\mathcal{R}_M$, conditions the spatial RIR on material information.
We follow a 4-layer transformer encoder~\cite{dosovitskiy2020vit} with a 256-dimensionality, feedforward of 512-dimensionality, and a dropout rate of 0.1.
We first use a convolutional patch extractor that patchifies each channel of the spatial RIR $\hat{A}_S$ separately.
We create patches of size $16\times16$ for each channel, then concatenate the tokens and use a 2-layer MLP to project such that an average feature is learned for each corresponding patch of the left and right channel spectrograms.
Furthermore, this module uses 4 re-weighting tokens \cite{hu2018squeeze} to learn cross-modality modulation between material and spatial acoustics.
These tokens are projected to the dimensionality of the encoder such that $R \in \mathbb{R}^{4\times256}$.
Features of $M, \hat{A}_S, R$ are concatenated and fed as input to $\mathcal{F}_M$, which performs self-attention across the input space. Next, we extract encoded audio tokens $g_m$ and re-weighting tokens $g_r$ from the encoder and feed them into their respective modules.

\vspace{-0.3cm}
\paragraph{Audio Feature Upsampling Network}
The upsampling network $\mathcal{U}_S$ consists of 4 transpose convolution layers.
Each layer consists of a transpose convolution with a kernel size of 2 and a stride of 2, followed by 2 convolutional layers and leaky ReLU activations~\cite{xu2015empirical}.
The input feature to the model is upsampled using [512, 256, 128, 64, 32] channels.
Finally, a two-layer network follows that maps the output to a 2-channel spectrogram of size $256 \times 256$.

\vspace{-0.3cm}
\paragraph{Material-Aware Audio Feature Upsampler}
Similar to $\mathcal{U}_S$, this module $\mathcal{U}_M$ upsamples feature map from the material encoder $g_m$.
Features from $g_r$ are independently projected to each layer of $\mathcal{U}_M$ using a linear projection followed by sigmoid ~\cite{han1995influence} to modulate the output in consecutive layers with cross-modal reweighting features.
The output of this module is the final estimation of the material-aware binaural spectrogram.

\vspace{-0.3cm}
\paragraph{Cross-Modal Correspondence Network}
This network uses dual ResNet18 architectures for encoding material map $M$ and predicted RIR $\hat{A}_M$, followed by a 3-layer MLP that classifies whether these two inputs correspond or not.
We pretrain this network on the training data from \cite{saad2025materialrir} for 10 epochs with Binary Cross-Entropy Loss, where 75\% of the batch, during training, consists of negative samples.
We achieve an 81\% accuracy on the unseen scene and unseen material split.
Once trained, we freeze the weights of this network and feed it with $M$ and predicted RIR $\hat{A}_M$ from $\mathcal{F}_M$ and use the output to provide feedback to the model during training.
We use Binary Cross-Entropy as an augmented training objective to the main model $\mathcal{F}$.

\subsection{Evaluation Setup and Baselines}\label{sec:supp_eval_setup}
We evaluate our approach against existing methods and baselines that cover different aspects of RIR generation.
\emph{Image2Reverb} and \emph{FAST-RIR++} are state-of-the-art methods for RIR prediction from visual and spatial cues.
However, both baselines perform poorly in our evaluations, even when trained on the AcoW dataset.
This suggests that relying only on visual or spatial cues is insufficient for generating accurate material-conditioned RIRs.

Our approach instead models spatial and material cues separately, producing spatially specific and materially specific RIRs for each scene and material assignment.
To isolate the effect of this disentangled design, we introduce \emph{JM-*} baselines that jointly model all environmental cues using the same input features as our method.
\emph{JM-CNN} uses a CNN decoder to estimate the material-conditioned RIR.
\emph{JM-Transformer} uses a transformer encoder, with the same architecture as our material encoder $\mathcal{R}_M$, to jointly encode all environmental features.
\emph{JM-QFormer} follows our spatial RIR decoder $\mathcal{R}_S$ and learns a set of tokens through cross-attention over the input environmental features.
We construct these baselines to study the effect of joint modeling and compare it with our disentangled RIR prediction approach.

Finally, we compare against the current state-of-the-art vision-based method for mapping material configurations to RIRs, M-CAPA.
This method uses the visual input $(V, M)$ to directly predict the material-conditioned RIR.
Our results show that separately predicting spatial and material RIRs yields more accurate results than state-of-the-art methods that model the material-conditioned RIR jointly.

\end{document}